\documentclass[runningheads]{llncs}
\usepackage{comment}
\usepackage{graphicx}
\usepackage{amsmath,amsfonts,bm}

\def\eqref#1{equation~\ref{#1}}
\def\1{\bm{1}}

\DeclareMathAlphabet{\mathsfit}{\encodingdefault}{\sfdefault}{m}{sl}
\SetMathAlphabet{\mathsfit}{bold}{\encodingdefault}{\sfdefault}{bx}{n}

\usepackage{hyperref}
\usepackage{url}
\usepackage{subfigure}
\usepackage{booktabs}
\usepackage{tablefootnote}
\usepackage{graphicx}
\usepackage{bm}
\usepackage{wrapfig}
\usepackage{changepage}

\usepackage{amsmath}
\usepackage{amsthm}
\usepackage{amssymb}
\usepackage{amsfonts,eucal,amsbsy}
\usepackage{mathtools}
\usepackage{multirow}
\usepackage{placeins}
\usepackage{float}
\usepackage{wrapfig}
\usepackage{subfigure}

\makeatletter
\newcommand{\printfnsymbol}[1]{%
  \textsuperscript{\@fnsymbol{#1}}%
}
\makeatother
\begin{document}
\title{Disentangled Contrastive Learning for Learning Robust Textual Representations}

\def\YOFOSubNumber{37}  % Insert your submission number here
%\begin{comment}
\titlerunning{CICAI2021 submission ID \YOFOSubNumber} 
\authorrunning{CICAI2021 submission ID \YOFOSubNumber} 
\author{Anonymous CICAI submission}
\institute{Paper ID 37}
%\end{comment}
%******************

 
\author{Xiang Chen\inst{1,2}\printfnsymbol{1}
\and Xin Xie \inst{1,2}\thanks{Equal contribution and shared co-first authorship.}
\and Zhen Bi \inst{1,2} 
\and Hongbin Ye \inst{1,2} 
\and Shumin Deng \inst{1,2}
\and Ningyu Zhang \inst{1,2}\printfnsymbol{2}
\and Huajun Chen \inst{1,2}\thanks{Corresponding author.}
}
\authorrunning{C. Author et al.}

\institute{Zhejiang University \& AZFT Joint Lab for Knowledge Engine, China \and
Hangzhou Innovation Center, Zhejiang University, China 
\email{\{xiang\_chen,xx2020,bi\_zhen,yehongbin,231sm,zhangningyu,huajunsir\}@zju.edu.cn}
}
 
%******************
\maketitle              % typeset the header of the contribution

\begin{abstract}
Although the self-supervised pre-training of transformer models has resulted in the revolutionizing of natural language processing (NLP) applications and the achievement of state-of-the-art results with regard to various benchmarks, this process is still vulnerable to small and imperceptible permutations originating from legitimate inputs. Intuitively, the representations should be similar in the feature space with subtle input permutations, while large variations occur with different meanings. This motivates us to investigate the learning of robust textual representation in a contrastive manner. However, it is non-trivial to obtain opposing semantic instances for textual samples. In this study, we propose a disentangled contrastive learning method that separately optimizes the uniformity and alignment of representations without negative sampling. Specifically, we introduce the concept of momentum representation consistency to align features and leverage power normalization while conforming the uniformity.  Our experimental results for the NLP benchmarks demonstrate that our approach can obtain better results compared with the baselines, as well as achieve promising improvements with invariance tests and adversarial attacks. The code is available in \url{https://github.com/zxlzr/DCL}.
%in \url{https://anonymous.4open.science/r/f82a2770-d44b-4e55-b849-4a85baaacde4/}
\end{abstract}

\section{Introduction}
The self-supervised pre-training of transformer models has revolutionized natural language processing (NLP) applications. Such pre-training with language modeling objectives provides a useful initial point for parameters that generalize well to new tasks with fine-tuning. However, there is a significant gap between task performance and model generalizability. Previous approaches have indicated that neural models suffer from poor \textbf{robustness} when encountering \emph{randomly permuted contexts} \cite{DBLP:conf/acl/RibeiroWGS20} and \emph{adversarial examples}  \cite{jin2019bert,li2021normal}. 

To address this issue, several studies have attempted to leverage data augmentation or adversarial training into pre-trained language models (LMs) \cite{jin2019bert}, which has indicated promising directions for the improvement of robust textual representation learning. Such methods generally augment textual samples with synonym permutations or back translation and fine-tune downstream tasks on those augmented datasets. Representations learned from instance augmentation approaches have demonstrated expressive power and contributed to the performance improvement of downstream tasks in robust settings. However, the previous augmentation approaches mainly focus on the supervised setting and neglect large amounts of unlabeled data. Moreover, it is still not well understood whether a robust representation has been achieved or if the leveraging of more training samples have contributed to the model robustness. 

Specifically, a robust representation should be similar in the feature space with subtle permutations, while large variations occur with different semantic meanings. This motivates us to investigate robust textual representation in a contrastive manner. It is intuitive to utilize data augmentation to generate positive and negative instances for learning robust textual representation via auxiliary contrastive objects. However, it is non-trivial to obtain opposite semantic instances for textual samples. For example, given the sentence, ``Obama was born in Honululu," we are able to retrieve a sentence such as, ``Obama was living in Honululu," or, ``Obama was born in Hawaii." There is no guarantee that these randomly retrieved sentences will have negative semantic meanings that contradict the original sample.  

In this study, we propose a novel disentangled contrastive learning (DCL) method for learning robust textual representations.  Specifically, we disentangle the contrastive object using two subtasks: feature alignment and feature uniformity \cite{wang2020understanding}. We introduce a unified model architecture to optimize these two sub-tasks jointly. As one component of this system, we introduce momentum representation consistency to align augmented and original representations, which explicitly shortens the distance between similar semantic features that contribute to feature alignment. As another component of this system, we leverage power normalization to enforce the unit quadratic mean for the activations, by which the scattering features within the same batch implicitly contribute to the feature uniformity. Our DCL approach is a unified, unsupervised, and model-agnostic approach, and therefore it is orthogonal to existing approaches. %We conduct numerous experiments on NLP benchmarks, which demonstrate the effectiveness of this approach in normal and robust settings. 
The contributions of this study can be summarized as follows:  

\begin{itemize}
\item  We investigate robust textual representation learning problems and introduce a disentangled contrastive learning approach. 

\item  We introduce a unified model architecture to optimize the sub-tasks of feature alignment and uniformity, as well as providing theoretical intuitions. 

\item  Extensive experimental results related to NLP benchmarks demonstrate the effectiveness of our method in the robust setting; we performed invariance tests and adversarial attacks and verified that our approach could enhance state-of-the-art pre-trained language model methods.
\end{itemize}

\section{Related  Work}

Recently, studies have shown that pre-trained models (PTMs) \cite{bert} on the large corpus are beneficial for downstream NLP tasks, such as in GLUE, SQuAD, and SNLI. The application scheme of these systems is to fine-tune the pre-trained model using the limited labeled data of specific target tasks.   Since training distributions often do not cover all of the test distributions, we would like a supervised classifier or model to perform well on. Therefore, a key challenge in NLP is learning robust textual representations.  Previous studies have explored the use of data augmentation and adversarial training to improve the robustness of pre-trained language models. %\cite{wei2019eda} proposed easy data augmentation techniques for boosting performance on text classification tasks. 
\cite{li2020textat} introduced a novel text adversarial training with token-level perturbation to improve the robustness of pre-trained language models. However,  supervised instance-level augmentation approaches ignore those unlabeled data and do not guarantee the occurrence of real robustness in the feature space.  
Our work is motivated by contrastive learning~\cite{pmlr-v97-saunshi19a}, which aims at maximizing the similarity between the encoded query $q$ and  matched key $k^{+}$, while distancing randomly sampled keys $\{k_0^{-},k_1^{-},k_2^{-},...\}$. 
By measuring similarity with a score function $s(q,k)$, a form of contrastive loss function is considered as:
\begin{equation}
    \mathcal{L}_{contrast} = -\log \frac{\exp(s(q,k^+))}{\exp(s(q,k^+)) + \sum_{i}\exp(s(q,k_{i}^-))},
\label{eq:ctl}
\end{equation}
where $k^{+}$ and $k^{-}$ are  positive and negative instances, respectively. The score function $s(q,k)$ is usually implemented with the cosine similarity $\frac{q^T k}{\parallel q \parallel \cdot \parallel k \parallel}$. $q$ and $k$ are often encoded by a learnable neural encoder (e.g., BERT~\cite{bert}).  Contrastive learning have increasingly attracted attention, which is beneficial for  unsupervised or self-supervised learning from computer vision~\cite{DBLP:conf/cvpr/WuXYL18,DBLP:conf/cvpr/YeZYC19,TianCMC,he2019Momentum,icml2020_6165} to natural language processing~\cite{ye2020contrastive,DBLP:conf/nips/MikolovSCCD13,DBLP:conf/nips/MnihK13,gunel2020supervised,wu2020clear,rethmeier2021primer}. 
%\cite{chi2020infoxlm} formulate cross-lingual language model pre-training as maximizing mutual information between multilingual-multi-granularity texts.
%\cite{DBLP:conf/iclr/ClarkLLM20} utilized a discriminator to predict whether a token is replaced by a generator given its surrounding context. 
%\cite{iter-etal-2020-pretraining} proposed to pre-train language models with contrastive sentence objectives to predict the surrounding sentences given an anchor sentence. 
%\cite{wei2020learning} proposed to encourage parallel cross-lingual sentences to obtain an identical semantic representation and distinguish whether a specific word is contained within these sentences. %To the best of our knowledge, this is the first study to apply contrastive learning to robust textual representation learning. 

\section{Preliminaries on Learning Robust Textual Representations}

\textbf{Definition 1.}\textbf{Robust textual representation} indicates that the representation is vulnerable  to small and imperceptible permutations originating from legitimate inputs. Formally, we have the following:
\begin{equation}
    g\left(X+z \right) = g(X), \text { and } \operatorname{Sim}\left(f(X+z), f(X)\right) \geq \epsilon,
\label{eq:ad_requirement}
\end{equation}

where $z$ refers to the random or adversarial permutation of the input text and $g(.)$ takes input from $x$ and outputs a valid probability distribution for tasks. $f(.)$ is the feature encoder, such as BERT. We are interested in deriving methods for pre-training representations that provide guarantees for the movement of inputs such that they are robust to permutations. Therefore, a robust representation should be similar in the feature space with subtle permutations, while large variations are observed for different semantic meanings. Such constraints are related to the well-known contrastive learning \cite{arora2019theoretical} schema as follows:

\textbf{Remark.}  Robust representation is closely related to regularizing the feature space with the following constraints: 
\begin{equation}
    L_{contrast}  = \sum (\sum_{1}^{m}|f(X) - f(X+z)| - \sum_{1}^{n}|f(X) - f(X^{\prime})|)
    \end{equation}
where $m$ and $n$ are the number of positive and negative instances, respectively, regarding the original input, $X$, $X+z$ and $X^{\prime}$ are the positive and negative instances, respectively. Note that we can obtain $X+z$ via off-the-shelf tools such as data augmentation or back-translation. However, it is non-trivial to obtain negative instances for textual samples. Previous approaches \cite{giorgi2020declutr,fang2020cert,chi2020infoxlm,wei2020learning} regard random sampling of the remaining instances from the corpus as negative instances; however, there is no guarantee that those random instances are semantically irrelevant. Recent semantic-based information retrieval approaches \cite{xiong2020approximate} can obtain numerous similar semantic sentences via an approximate nearest neighbor \cite{liu2005investigation}, which further indicates that negative sampling for sentences may result in noise.   
In this study, inspired by the approach utilized by \cite{wang2020understanding}, we 
disentangle the contrast loss with the two following properties: \begin{itemize}
    \item \emph{Alignment}: two samples forming a positive pair should be mapped to nearby features and therefore be (mostly) invariant to unneeded noise factors.
    \item \emph{Uniformity}: feature vectors should have an approximately uniform distribution on the unit hypersphere. %, thereby preserving as much information of the data as possible.
\end{itemize}
\begin{equation}\begin{split}
L_{\text {contrast }}=&\mathbb{E}\left[-\log \frac{e^{f_{x}^{T} f_{y} / \tau}}{e^{f_{x}^{T} f_{y} / \tau}+\sum_{i} e^{f_{x}^{T} f_{y_{i}^{-} / \tau}}}\right] \\
=&\mathbb{E}\left[-f_{x}^{T} f_{y} / \tau\right]+\mathbb{E}\left[\log \left(e^{f_{x}^{T} f_{y} / \tau}+\sum_{i} e^{f_{z}^{T} f_{y_{i}^{-} / \tau}}\right)\right] \\
\overset{\left.\mathbf{P}\left[f_{, v}=f_{y}\right)\right]=\mathbf{1}}{=}& \underbrace{\mathbb{E}\left[-f_{x}^{T} f_{y} / \tau\right]}_{\text {positive alignment }}+ \underbrace{\mathbb{E}\left[\log \left(e^{1 / \tau}+\sum_{i} e^{f_{x}^{T} f_{y_{i}^{-} / \tau}}\right)\right]}_{\text {uniformity}}
\end{split}
\end{equation}
The alignment loss can be defined straightforwardly as follows:  %
\begin{equation}
\mathcal{L}_{\text {align }}(f ; \alpha) \triangleq-\underset{(x, y) \sim p_{\text {pos }}}{\mathbb{E}}\left[\|f(x)-f(y)\|_{2}^{\alpha}\right], \quad \alpha>0
\end{equation}
Where $f(.)$ is the feature encoder and $x$,$y$ are positive instance pairs.  The uniformity metric refers to  optimizing this metric should converge to a uniform distribution. %Note that feature uniformity should be  empirically reasonable with a finite number of points and asymptotically correct. Therefore, 
The loss can be defined with the radial basis function (RBF) kernel $G_{t}: \mathcal{S}^{d} \times \mathcal{S}^{d} \rightarrow \mathbb{R}_{+}$ \cite{wang2020understanding}. Formally, we have: 
\begin{equation}\begin{split}
    L_{uniform}(f; t)
    &  \triangleq \log \underset{x, y \stackrel{\mathrm{ii.d.}}{\mathbb{E}}}{\mathbb{E}}\left[G_{t}(u, v)\right] \\
    & = \log \underset{x, y^{\mathrm{id.d.}} p_{\text {data }}}{\mathbb{E}}\left[e^{-t\|f(x)-f(y)\|_{2}^{2}}\right], \quad t>0
\end{split}\label{eq_uniform}
\end{equation}
where $t$ is a fixed parameter.

\section{Disentangled Contrastive Learning}

 \begin{figure*}
  \centering
  \includegraphics[width=1\textwidth]{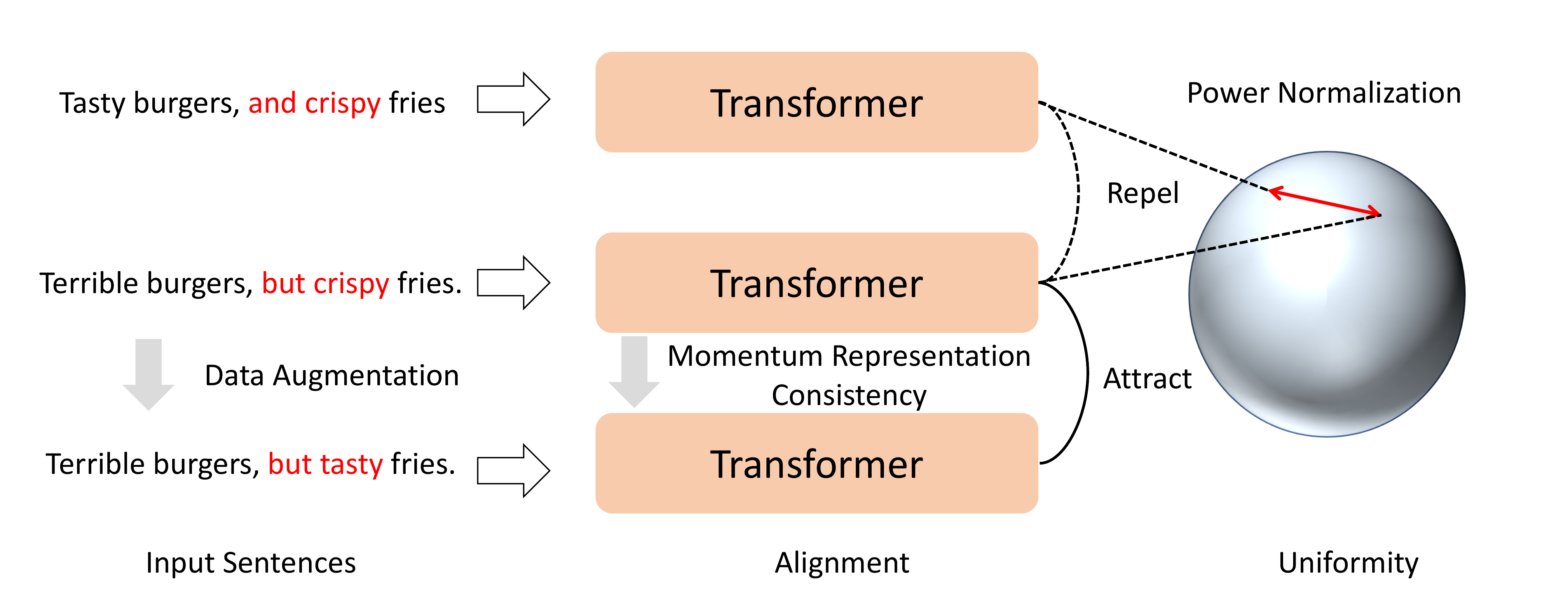}
  \caption{Disentangled contrastive learning for robust textual representations.}\label{arc}
\end{figure*}
 
In this section, we present a preliminary study on how to learn robust textual representation via disentangled contrastive learning, as represented in Figure \ref{arc}. %Because the aforementioned analysis shows that contrastive learning can be disentangled with feature alignment and uniformity, it is intuitive to optimize the representation learning method with separated objects, thereby learning without negative textual instances.  

\subsection{Feature Alignment with Momentum Representation Consistency} 

There are multiple ways to align a textual representation. We utilize two transformers with a consistent momentum representation to explicitly guarantee feature alignment \cite{grill2020bootstrap}. The two networks are defined by a set of weights $\theta$ and $\xi$. We use the exponential moving average of the parameters $\theta$ to get $\xi$. Formally, we have:
\begin{equation}
    \xi \leftarrow \tau \xi+(1-\tau) \theta
\end{equation}
Given a sentence $\mathcal{X}$ and its augmentation $\mathcal{X}^{\prime}$ (e.g, via data augmentation) from the first original network, we may obtain  output representations $q \triangleq f_{\theta}(X)$ and  $p \triangleq f_{\theta}(X^{\prime})$. Note that previous works \cite{icml2020_6165,grill2020bootstrap} indicates that an projection $p$ in feature space improve the performance. We then leverage a projection function $g(p_{\theta})$ and $\ell_2$-normalize both $g(p_{\theta})$ and $q_{\xi}$ to $\bar{g}(p_{\theta}) \triangleq g/\|g(p_{\theta})\|_2$ and $\bar{q}_{\xi} \triangleq q_{\xi} / \|q_{\xi}\|_2$, respectively.  We  leverage the mean squared loss as follows:
\begin{equation}
\mathcal{L}_{\mathrm{align}} \triangleq\left\|\overline{g}\left(q\right)-\bar{p}_{\xi}\right\|_{2}^{2}=2-2 \cdot \frac{\left\langle g\left(q_{\theta}\right), p_{\xi}\right\rangle}{\left\|g\left(q_{\theta}\right)\right\|_{2} \cdot\left\|p_{\xi}\right\|_{2}}
\end{equation}

Additionally, we make the losses symmetrical $\mathcal{L}_{\mathrm{align}}$ by  feeding $X$ to the augmented network and $X^{\prime}$, separately. We optimize  $\mathcal{L}_{\mathrm{align}}+\widetilde{\mathcal{L}}_{\mathrm{align}}$ with respect to $\theta$ only, but \emph{not} $\xi$, via the stop-gradient.  
 
\subsection{Feature Uniformity with Power Normalization} 

To ensure that feature vectors should have an approximately uniform distribution, we can directly optimize the Eq. \ref{eq_uniform}. However, different from computer vision, in the original loss of BRET \cite{bert}, we have already utilized the next sentence prediction loss. Such a contrastive object has explicitly made the sentence representation $f(.)$ scattered in the feature space; thus, the model may quickly collapse without learning. Inspired by \cite{santurkar2018does}, we argue that batch normalization can identify the common-mode between examples of a mini-batch and removes it using the other representations in the mini-batch as implicit negative examples. We can, therefore, view batch normalization as a novel method of implementing feature uniformity on embedded representations. Because vanilla batch normalization will lead to significant performance degradation when naively used in NLP, we leverage an enhanced power normalization \cite{shenpowernorm} to guarantee feature uniformity. Specifically, we leverage the unit quadratic mean rather than the mean/variance of running statistics with an approximate backpropagation method to compute the corresponding gradient. Formally, we have the following:

\begin{equation}
    \begin{aligned} \widehat{\boldsymbol{X}}^{(t)} &=\frac{\boldsymbol{X}^{(t)}}{\psi^{(t-1)}} \\ \boldsymbol{Y}^{(t)} &=\gamma \odot \widehat{\boldsymbol{X}}^{(t)}+\beta \\\left(\psi^{(t)}\right)^{2} &=\left(\psi^{(t-1)}\right)^{2}+(1-\alpha)\left(\psi_{B}^{2}-\left(\psi^{(t-1)}\right)^{2}\right) \end{aligned}
\end{equation}
Note that we compute the gradient of the loss regarding the quadratic mean of the batch. In other words, we utilize the running statistics to conduct backpropagation, thus, resulting in bounded gradients, which is necessary for convergence in NLP (see proofs in \cite{shenpowernorm}). 

\subsection{Implementation Details}
We leverage synonyms from  WordNet categories to conduct data augmentation for computation efficiency.  We combine all the momentum representation consistency and power normalization results in a unified architecture with the mask language model object. We leverage the same architecture of the BERT-base \cite{bert}. We first pre-train the model in a large-scale corpus unsupervisedly (e.g., the same corpus and training steps with BERT) and then fine-tune the model using task datasets. 

\section{Experiment} 
 
%We evaluated our method using NLP benchmarks, including tasks of text classification, natural language inference, machine reading comprehension, and the GLUE series of language understanding tasks. We conduct experiments on the normal test set as well as robust settings (e.g., invariance tests and adversarial attacks). The code and datasets are available at \url{anonymous}.

\subsection{Datasets and Setting}

\begin{table*}[ht]
\setlength{\textwidth}{4.5pt}
\small
    \centering
    \caption{Summary of results on GLUE.} \label{tbl:results1}
    \begin{tabular}{ll|ccccccc|c}
     \toprule
      \multicolumn{2}{c|}{\multirow{2}{*}{\textbf{Model}}} & \multirow{2}{*}{\textbf{\textsc{CoLA}}} & \multirow{2}{*}{\textbf{\textsc{SST-2}}} & \multirow{2}{*}{\textbf{\textsc{MRPC}}} & \multirow{2}{*}{\textbf{\textsc{QQP}}} &
      \textbf{\textsc{MNLI}} & \multirow{2}{*}{\textbf{\textsc{QNLI}}} & \multirow{2}{*}{\textbf{\textsc{RTE}}} &
      \textbf{\textsc{GLUE}} \\
      &&&&&& \textbf{\textsc{(m/mm)}} & & & \textbf{\textsc{Avg}}\\
      \midrule
      \multirow{3}{*}{\rotatebox[origin=c]{0}{\textsc{Normal}}} & BERT &56.8  &92.3  &89.7  &89.6  &84.6/85.2 &91.5  &69.3  & 82.3 \\
            & BERT+DA &58.6  &93.2  &86.5  &86.7  &84.2/84.4 &91.1  &68.9  & 81.7 \\
            & \textsc{DCL} &\textbf{60.9}  &93.0  &\textbf{89.7}  &\textbf{90.0}  &84.7/84.6 &\textbf{91.7}  &\textbf{69.7}  & \textbf{83.0} \\
      \midrule
      \multirow{3}{*}{\rotatebox[origin=c]{0}{\textsc{Robust}}} &BERT &46.4   &91.8  &88.1  &84.9  &81.6/82.2 &89.2  &67.1  & 78.9\\
         & BERT+DA &53.8  &92.9  &85.6  &85.5   & 83.1/83.4 &90.7  &66.3  & 80.1 \\
            & \textsc{DCL} &48.4  &\textbf{92.4}  &86.0  &\textbf{85.5}  &82.5/82.7   &89.7  &\textbf{68.8}  & 79.5 \\
      \bottomrule
    \end{tabular}
    %\vspace{-0.5cm}
    \label{main}
\end{table*}

We conducted experiments on three benchmarks: GLUE, SQuAD, SNLI, and DialogRE. 

\textbf{GLUE} \cite{wang2019glue} is an NLP benchmark aimed at evaluating the performance of downstream tasks of the pre-trained models. Notably, we leverage nine tasks in GLUE, including CoLA, RTE, MRPC, STS, SST, QNLI, QQP, and MNLI-m/mm. We follow the same setup as the original BERT for single sentence and sentence pair classification tasks. We leverage a multi-layer perception with a softmax layer to obtain the predictions. 

\textbf{SQuAD} is a reading comprehension dataset constructed from Wikipedia articles. We report results on SQuAD 1.1. Here also, we follow the same setup as the original BERT model and predict an answer span---the start and end indices of the correct answer in the correct context. 

\textbf{SNLI} is a collection of 570k human-written English sentence pairs that have been manually labeled for balanced classification with entailment, contradiction, and neutral labels, thereby supporting the task of natural language inference (NLI). We add a linear transformation and a softmax layer to predict the correct label of NLI. 

\textbf{DialogRE} is a dialogue-based relation extraction dataset, which contains 1,788 dialogues from a famous American television situation comedy Friends.

To evaluate the robustness of our approach, we also conduct invariance testing with CheckList\footnote{\url{https://github.com/marcotcr/checklist.git}} \cite{DBLP:conf/acl/RibeiroWGS20} and
adversarial attacks\footnote{\url{https://github.com/thunlp/OpenAttack}}. To generate label-preserving perturbations, we used WordNet categories (e.g., synonyms and antonyms). We selected context-appropriate synonyms as permutation candidates. To generate adversarial samples, we leverage a probability-weighted word saliency (PWWS) \cite{ren2019generating} method based on synonym replacement. We manually evaluate the quality of the generated instances. We also conduct experiments that apply data augmentation and adversarial training to the BERT model. 
We utilize PyTorch to implement our model. We use Adam optimizer with a cosine decay learning rate schedule.  We set the initial learning rate as 1e-5. We use a batch size of 32 over eight Nvidia 1080Ti GPUs. %With this setup, training takes approximately one month. We leverage the grid search to find optimal hyper-parameters in the development set. We ran each experiment five times and calculated the average performance. 

\subsection{Results  and Analysis}

%\vspace{0.5cm}
\textbf{Main Results}

\begin{wraptable}{r}{0.4\textwidth}
    \centering
    \caption{Summary of results on SQuAD.} \label{tbl:results2}
    \begin{tabular}{ll|cc}
     \toprule
\multicolumn{2}{c|}{\textbf{Model}} & \textbf{F1} & \textbf{EM} \\
      \midrule
      \multirow{3}{*}{\rotatebox[origin=c]{0}{\textsc{Normal}}} & BERT & 88.5 & 80.8 \\
      &BERT+DA&  88.2&  80.4  \\
      & \textsc{DCL} &   \textbf{88.4}&  \textbf{81.0}    \\
      \midrule
      \multirow{3}{*}{\rotatebox[origin=c]{0}{\textsc{Robust}}} &BERT & 86.7 &  77.8 \\
      &BERT+DA &87.8  &79.9   \\
      & \textsc{DCL} &86.8  &78.1    \\
      \bottomrule
    \end{tabular}
    \vspace{-0.5cm} 
    \label{squad}
\end{wraptable}

From Table \ref{main} and \ref{squad}, we can observe the following: 
1)  Vanilla BERT achieves poor performance in the robust set on both GLUE and SQUAD, which indicates that the previous fine-tuning approach cannot obtain a robust textual representation. This will lead to performance decay with permutations.

2)  With data augmentation, BERT can obtain improved performance in the robust set; however, a slight performance decay is observed in the original test set. We argue that data augmentation can obtain better performance by fitting to task-specific data distribution; there is no guarantee that more data will result in robust textual representations.

3)  Our DCL approach achieves improved performance in both the original test set and robust set compared with vanilla BERT. Note that our DCL is an unsupervised approach, and we leverage the same training instances with BERT. The performance improvements indicate that our approach can obtain more robust textual representations that enhance the performance of the system.

\textbf{Adversarial Attack Results}

\iffalse
\begin{wraptable}{r}{0.7\textwidth}
\setlength{\tabcolsep}{4.5pt}
\small
    \centering
    \caption{Summary of results on CoLA, SNLI, DialogRE.} \label{tbl:results2}
    \begin{tabular}{ll|c|c|c}
     \toprule
\multicolumn{2}{c|}{\textbf{Model}} & \textbf{CoLA} & \textbf{SNLI} & \textbf{DialogRE} \\
      \midrule
      \multirow{3}{*}{\rotatebox[origin=c]{0}{\textsc{Normal}}} & BERT &56.8  &91.0 &63.0 \\
      &BERT+Adv &55.0   &90.9 & 64.3\\
      & \textsc{DCL} &\textbf58.8    &91.0&  64.2\\
      \midrule
      \multirow{3}{*}{\rotatebox[origin=c]{0}{\textsc{Adversarial}}} &BERT &47.0 &87.4  &62.0\\
      &BERT+Adv &55.1  &90.3 & 62.9\\
      & \textsc{DCL} &48.2   &90.5&  63.2\\
      \bottomrule
    \end{tabular}
    \vspace{-0.5cm} \label{adv}
\end{wraptable}
\fi

\begin{wraptable}{r}{0.7\textwidth}
\setlength{\tabcolsep}{4.5pt}
\small
    \centering
    \caption{Summary of results on CoLA, SNLI, DialogRE.} \label{tbl:results2}
    \begin{tabular}{ll|c|c|c}
     \toprule
\multicolumn{2}{c|}{\textbf{Model}} & \textbf{CoLA} & \textbf{SNLI} & \textbf{DialogRE} \\
      \midrule
      \multirow{3}{*}{\rotatebox[origin=c]{0}{\textsc{Normal}}} & BERT &56.8  &91.0 &63.0 \\
      &BERT+Adv &55.0   &90.9 & 64.3\\
      & \textsc{DCL} &\textbf58.8    &91.0&  64.2\\
      \midrule
      \multirow{3}{*}{\rotatebox[origin=c]{0}{\textsc{Adversarial}}} &BERT &47.0 &87.4  &59.0\\
      &BERT+Adv &55.1  &90.3 & 62.9\\
      & \textsc{DCL} &48.2   &90.5&  63.2\\
      \bottomrule
    \end{tabular}
    \vspace{-0.5cm} \label{adv}
\end{wraptable}

From Table \ref{adv}, we can observe the following: 
1)  Vanilla BERT achieves a poor performance with adversarial attacks; BERT with adversarial training can obtain a good performance. However, we notice that there exists a performance decay for adversarial training in the original test set. Note that adversarial training methods would lead to standard performance degradation \cite{wen2019towards}, i.e., the degradation of natural examples. 
2) Our DCL approach achieves improved performance in the test set with and without an adversarial attack, which further demonstrates that our approach can obtain robust textual representations that are stable for different types of permutations. 

\textbf{Quantitative Analysis of Textual Representation}

\begin{figure*}
\centering
\subfigure[BERT(Random)] { \label{aa}
  \includegraphics[width=0.22\textwidth]{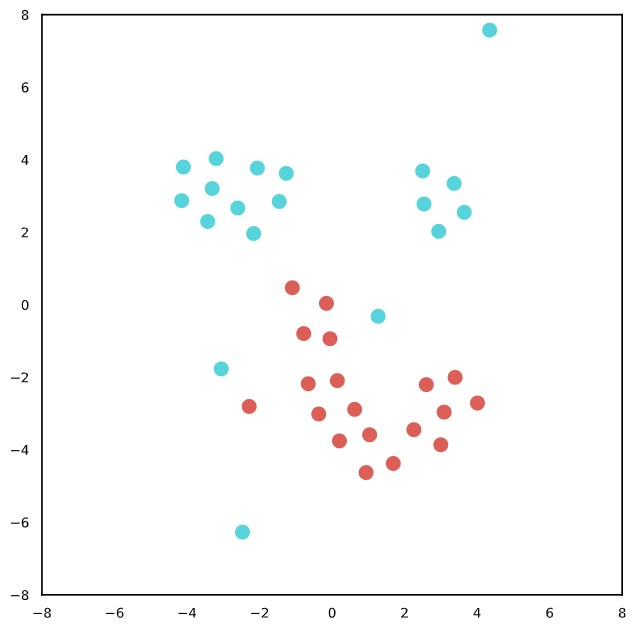}
}
\subfigure[DCL(Random)] { \label{cc}
\includegraphics[width=0.22\textwidth]{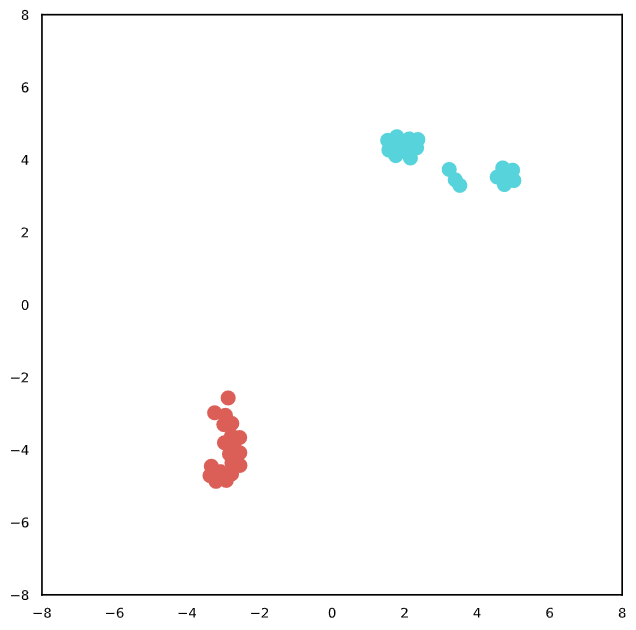}
}
\subfigure[BERT(Adv)] { \label{bb}
\includegraphics[width=0.22\textwidth]{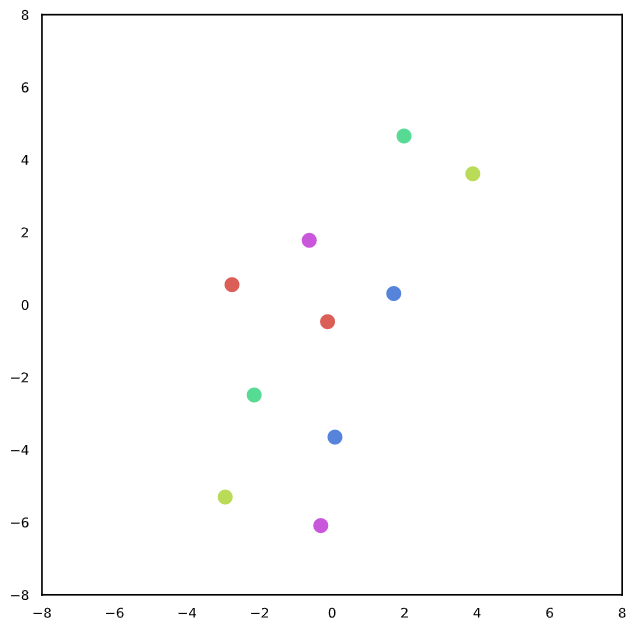}
}
\subfigure[DCL(Adv)] { \label{dd}
\includegraphics[width=0.22\textwidth]{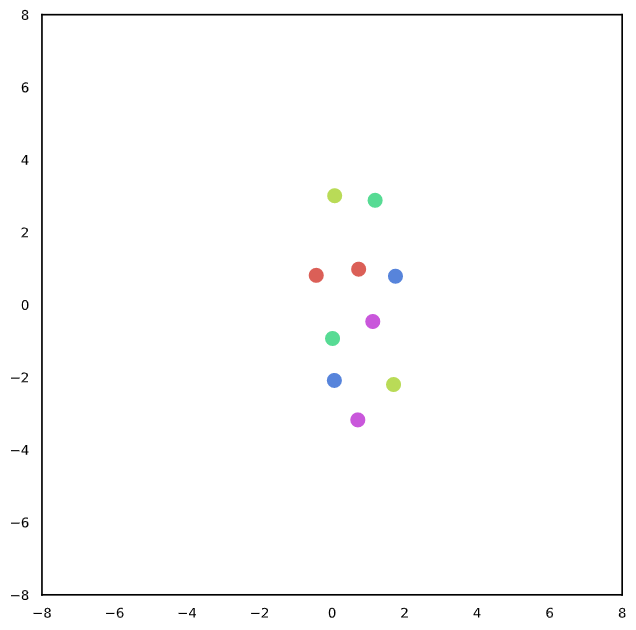}
}
\caption{T-SNE visualizations of sentence embeddings.}
\label{vis}
\end{figure*}

As we hypothesize that power normalization can implicitly contribute to feature uniformity, we conduct further experiments to analyze the effects of normalization \cite{2020blog}. Specifically, we random sample instances and leverage the cosine similarity of the original input projection vectors and the augmented projection vectors. We calculate the average cosine similarity between positive instances (in blue) and random instances (in red) with different strategies, including without normalization (No Norm), batch normalization (BN), and power normalization.  

\begin{wrapfigure}{r}{0.4\textwidth}
  \centering
  \includegraphics[width=0.4\textwidth]{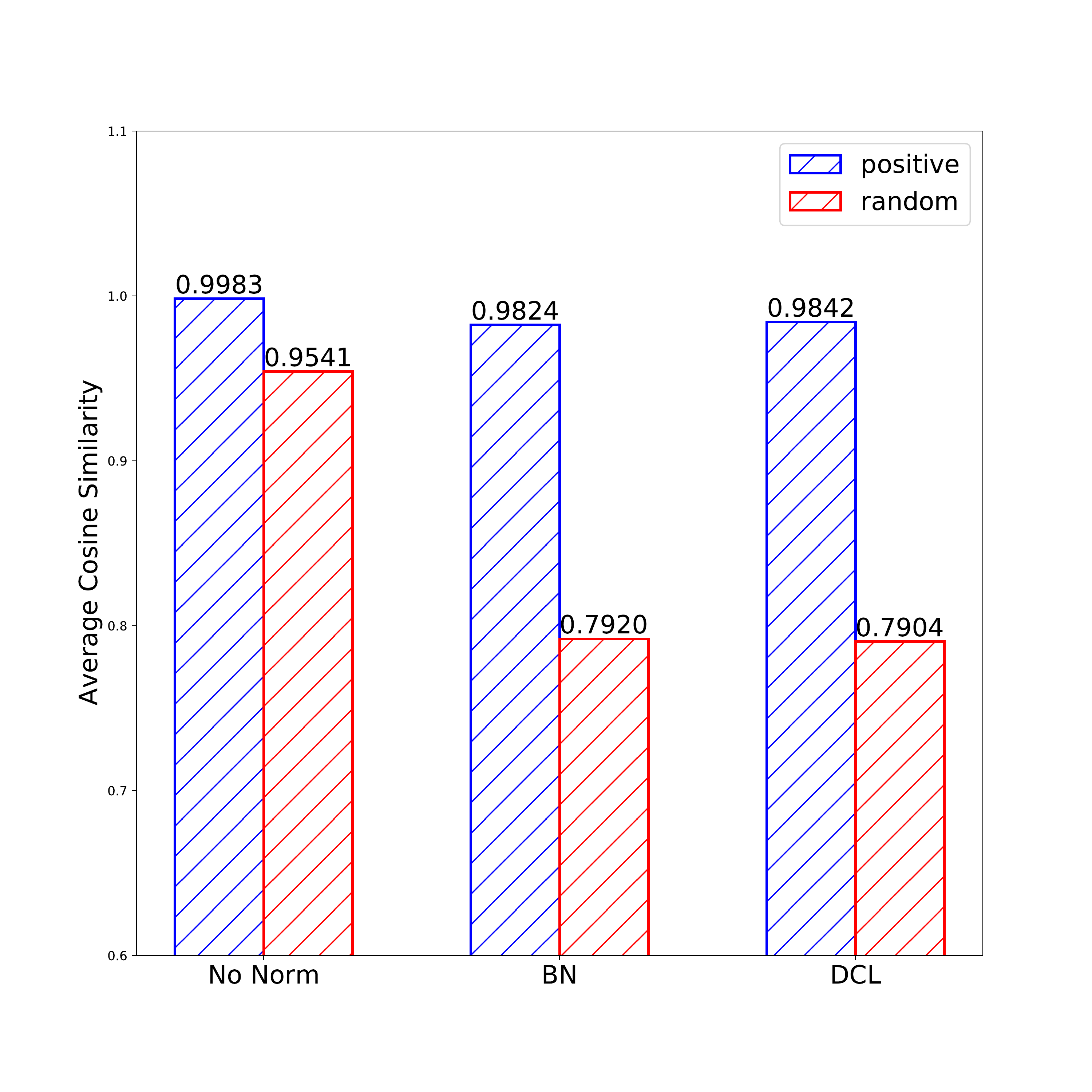}
  \caption{Cosine similarity of the original input projection vectors with the augmented input projection vectors.}\label{bn}
\end{wrapfigure}

From Figure \ref{bn}, we observe that with no normalization in $p$ or $q$, the feature space is aligned for both positive and negative instances, which shows that there exists a feature collapse for textual representation learning.  Considering  DCL training (i.e., with power normalization), we notice that the textual representations are relatively more similar between the positive instances (0.9842) than random (negative) ones (0.7904); thus, we can obtain different vectors.

Next, we give an intuitive explanation of preventing feature collapse for textual representation learning. Given an input instance without negative examples, the model may always output the projection vector $z$ with $[0, 1, 0, 0, …]$. Thus, the model can achieve a perfect prediction through learning a simple identity function, which, in other words, collapse in the feature space. With normalization, the output vector $z$ cannot obtain such singular values. Since the outputs will be redistributed regarding the learned mean and standard deviation, we can implicitly learn robust representations. 

\textbf{Qualitative Analysis of Textual Representation}

We randomly selected instances to visualize a sentence with T-SNE  \cite{maaten2008visualizing} to better understand the behaviors of textual representations. The different color refers to the different sentence pairs for both random permutation and adversarial attack settings. From Figure \ref{vis}, it may be observed that our approach can obtain a relatively similar semantic representation with permutations in both invariant tests and adversarial attack settings. Note that we explicitly align the projection of the textual representation with a random permutation, thereby encouraging similar semantic instances to have relatively similar representations.  

\subsection{Discussion}

\textbf{Robust Representation with Contrastive Learning.} Conventional approaches usually try to leverage instance-level augmentation aimed at achieving good performance on a robust set. However, there is no guarantee that robust textual representations will be obtained. Intuitively, directly aligning the representation of input tokens with slight permutations may contribute to robust representations. However, without any negative constraints, the model will easily collapse with a sub-optimal solution.  In this study, we observe that power normalization identifies this common mode between examples. In other words, it can remove those trivial samples by using the other representations in the batch as implicit negative instances. We can, therefore, view normalization as an implicitly contrastive learning method. 

\textbf{Limitations.} This work is not without limitations. We only consider the synonym replacement as a data augmentation strategy due to the efficiency of processing a huge amount of data. Other strong data augmentation methods can also be leveraged. Another issue is representation alignment, as there are lots of augmentations. We cannot enumerate all positive pairs for alignments; thus, there is still some room for designing more efficient feature-aligning algorithms. Moreover, as we utilize the square root loss, which is absolutely a Euclidean distance. Recent approaches \cite{meng2020hierarchical} indicates that Euclidean space may be sub-optimal for textual representations, and we leave this for future works.

%Lastly, with power normalization, the network outputs are no longer learning a pure function of the corresponding inputs. Thus,  it may be interesting to develop methods to avoid the use of power normalization during training. %Moreover, it may be promising to investigate alternative methods, such as weight standardization with group normalization for textual representation learning. 

\section{Conclusion}
We investigated robust textual representation learning and proposed a disentangled contrastive learning approach. We introduced feature alignment with a momentum representation consistency and feature uniformity with power normalization. 
We empirically observed that our approach could obtain an improved performance compared with baselines in NLP benchmarks and achieve a robust performance with invariant tests and adversarial attacks. 
%We also performed quantitative and qualitative analyses for learned textual representations, which indicated that our approach mitigates model collapse and can learn robust textual representations. %This may provide a basis for future works concerning robust representation learning. Our approach is model-agnostic; therefore, it can be applied to any pre-trained language models.  
%Further research on robust textual representation learning may be conducted to investigate such topics as 1) exploiting multi-task learning for robust representations; 2) investigating the essence of model robustness and proposing more efficient approaches to learn robust representations; and 3) incorporating more complex views (e.g., higher-order or skip n-grams, syntactic and semantic parses, etc.) and designing appropriate self-supervised tasks. 

\section{Acknowledgments}
We  want to express gratitude to the anonymous reviewers for their hard work and kind comments. This work is funded by NSFC91846204/NSFCU19B2027.

\bibliographystyle{splncs04}
\bibliography{icaai}

\end{document}